\documentclass[a4paper, 11pt]{article}
\usepackage[utf8]{inputenc}
\usepackage{graphicx}
\usepackage{amsmath}
\usepackage{braket}
\usepackage{caption}
\usepackage{subcaption}
\usepackage{bbold}
\usepackage{amssymb}
\usepackage[margin=1in]{geometry}
\usepackage{mathtools}
\usepackage{hyperref}
\usepackage{booktabs}
\usepackage{multirow}
\usepackage{authblk}
\usepackage{appendix}
\usepackage{algorithm}
\usepackage{algpseudocode}
\usepackage[backend=biber,sorting=none,style=ieee,natbib]{biblatex}
\usepackage{ifthen}
\usepackage{tikz}
\usepackage{amsmath}
\usepackage{booktabs}
\usepackage{multirow}
\usepackage{colortbl}
\hypersetup{
    linktocpage=true,
    colorlinks=true,
    linkcolor=blue,
    citecolor=blue,
}
\DeclareUnicodeCharacter{2212}{-}
\title{\textbf{Evaluating the Impact of Compression Techniques on Task-Specific Performance of Large Language Models}}

\author[1]{Bishwash Khanal}
\author[2]{Jeffery M. Capone}
\affil[1]{\href{bishwash.khanal@optiml.org}{\texttt{bishwash.khanal@optiml.org}}}
\affil[2]{\href{jeff.capone@optiml.org}{\texttt{jeff.capone@optiml.org}}}
\affil[ ]{OptiML Org, California, USA}
\date{}

\begin{document}
\maketitle
\providecommand{\keywords}[1]
{
  \small	
  \textbf{\textit{Keywords:}} \textit{#1}
}

\begin{abstract}
  Large language models (LLMs) offer powerful capabilities but incur substantial computational costs, driving the need for efficient compression techniques. 
  This study evaluates the impact of popular compression methods - Magnitude Pruning, SparseGPT, and Wanda - on the LLaMA-2-7B model, focusing on the trade-offs 
  between model size reduction, downstream task performance, and the role of calibration data. Our findings reveal that while SparseGPT and Wanda preserve 
  perplexity even at 50\% sparsity, they suffer significant degradation on downstream tasks, highlighting the inadequacy of perplexity as the sole evaluation 
  metric. To address this, we introduce Jensen-Shannon (JS) Divergence as a more comprehensive metric that captures nuanced changes in model behavior 
  post-compression. We further demonstrate that task-specific calibration data significantly enhances the downstream performance of compressed models compared 
  to general calibration data. This research underscores the necessity for diverse evaluation metrics and careful calibration data selection to fully 
  understand the complexities of LLM compression and its implications for practical applications.

\end{abstract}

\section{Introduction} \label{sec:intro}
Large language models (LLMs) like GPT-4 \cite{openai2024gpt4technicalreport}, PaLM \cite{10.5555/3648699.3648939}, and LLaMA \cite{llama1,llama2,llama3} 
have demonstrated remarkable capabilities across multi-task language understanding. However, the immense size of these models, often consisting of billions 
of parameters, presents significant challenges in computational requirements, memory footprint, and energy consumption during both training and inference. 
Consequently, there is growing interest in developing compression techniques to mitigate these costs while retaining model performance.

Several compression approaches, such as pruning, quantization, and knowledge distillation, have been proposed \cite{wang2024modelcompressionefficientinference}. 
Network pruning aims to shrink network sizes by removing specific weights from the model by setting them to zero. Weight quantization is the process of 
quantizing model parameters into lower bit-level representations. And knowledge distillation involves training a smaller, student model to mimic the 
behavior of a larger, teacher model. This process transfers the knowledge from the teacher to the student, achieving a more compact and efficient model 
without significant loss in performance. Although promising from a traditional model evaluation metric perspective, the impact of these methods on the  
performance of compressed models on downstream tasks remains an area of active research \cite{jaiswal2024compressingllmstruthrarely}.

Traditional metrics like perplexity do not fully capture the effects of compression on task-specific performance, necessitating the use of alternative metrics 
\cite{jaiswal2024compressingllmstruthrarely}. To address this limitation, we propose using Jensen-Shannon Divergence (JS) \cite{61115} between the base model 
and the compressed model on random data. This metric provides a better sense of how much the model has changed due to compression, offering insights into the 
model's capability to maintain performance on specific tasks. We explore its application in assessing the impact of compression on downstream performance.

Furthermore, effective compression techniques often leverage calibration data \cite{wang2024modelcompressionefficientinference}, making it essential to 
understand how this data influences downstream performance. This ensures that compressed models remain effective for their intended tasks. Our study, therefore, not only introduces 
JS Divergence as an alternative metric but also examines the critical role of calibration data in shaping the performance of compressed models on downstream tasks.

\section{Related Works} \label{sec:literature}

Current research has focused on compressing large language models (LLMs) while maintaining their broad, general capabilities without 
sacrificing accuracy, typically measured by perplexity \cite{frantar2023sparsegptmassivelanguagemodels,sun2024simpleeffectivepruningapproach, gptq, awq}. 
Perplexity, a standard measure of a model's performance, evaluates how confidently the model predicts the next token in a sequence. Despite the vast 
number of possible tokens, the model effectively narrows its prediction to just 2 likely candidates, demonstrating its efficiency in 
reducing uncertainty and making accurate predictions. Perplexity is expressed as:

\begin{equation}
    \text{Perplexity}(P) = 2^{H}
\end{equation}
\begin{equation}
    H = - \frac{1}{N} \sum_{i=1}^{N} \log_{2}(P(w_{i}|w_{1},w_{2},...,w_{i-1}))
\end{equation}

where $H$ is the average cross-entropy across $N$ tokens in the sequence, and $w_i$ represents the $i$-th token in the sequence. While perplexity is useful for evaluating the general capability of a model, 
recent attention has shifted to assessing whether claims of significant size reductions with negligible accuracy loss hold for specific downstream 
tasks such as question answering, code generation, and instruction-following. Findings indicate that sparsified models often underperform on 
these specialized tasks \cite{jaiswal2024compressingllmstruthrarely}. However, the Lottery Ticket Hypothesis \cite{frankle2019lotterytickethypothesisfinding} 
has shown that strategically designed sparse networks can match or even outperform dense counterparts when fine-tuned for specific tasks. 
This insight motivates our study on the impact of calibration data for compressing models using SparseGPT \cite{frantar2023sparsegptmassivelanguagemodels} 
and Wanda \cite{sun2024simpleeffectivepruningapproach}.

Despite the widespread use of perplexity, its limitations as a sole metric for evaluating LLMs have been increasingly recognized. 
For instance, \cite{Muhlgay2023GeneratingBF} highlights that perplexity may not adequately capture a model's ability to handle factual knowledge, 
as it does not contrast facts with false statements. This is further supported by \cite{hu2024can,sourabh2024ppl}, which discusses the challenges 
of using perplexity, such as its emphasis on immediate contextual prediction over broad understanding, difficulties in capturing 
ambiguity and creativity, and the impact of vocabulary on model performance.

In response to these limitations, alternative evaluation methods have been proposed. The OpenLLM Leaderboard \cite{myrzakhan2024openllm} is an 
open-source project aimed at tracking and evaluating open-sourced LLMs and chatbots, providing a more comprehensive assessment 
of model performance across diverse benchmarks. The Chatbot Arena \cite{chiang2024chatbot} is another open platform 
for evaluating LLMs through crowdsourced, pairwise human preferences, providing valuable insights into model performance from a 
human perspective. Additionally, \cite{NEURIPS2023_91f18a12} explores the use of LLMs as evaluators, finding that strong LLM judges like GPT-4 can 
match both controlled and crowdsourced human preferences well.

Furthermore, \cite{kaddour2023} establishes a systematic set of open problems and application successes for LLMs, helping ML researchers 
comprehend the field's current state more quickly and become productive. 

While perplexity has been a standard metric for evaluating LLMs, its limitations in capturing the full impact of model compression have been increasingly recognized. 
Previous studies \cite{jaiswal2024compressingllmstruthrarely,Muhlgay2023GeneratingBF,hu2024can,sourabh2024ppl} have pointed out these shortcomings but have not proposed a comprehensive alternative. 
This study addresses this issue by introducing Jensen-Shannon (JS) Divergence as a more holistic metric, providing deeper insights into the effects of compression on downstream task performance. 
By aligning JS Divergence with GPT-4 evaluations, this work offers a practical and cost-effective method for assessing compressed models, thereby advancing the field of LLM compression.
\section{JS Divergence as a Evaluation Metric} \label{sec:methodology} 

To evaluate the impact of compression techniques on task-specific performance in large language models, we selected the LLaMA-2-7B model 
\cite{llama2} due to its balance between efficiency and complexity. Despite its lower compressibility compared to its predecessor Open Pre-trained 
Transformer (OPT) models \cite{zhang2022opt}, it provides a rigorous test for evaluating the effectiveness of different compression methods.

We employed three popular compression techniques: Magnitude Pruning \cite{10.5555/2969239.2969366}, SparseGPT \cite{frantar2023sparsegptmassivelanguagemodels}, 
and Wanda \cite{sun2024simpleeffectivepruningapproach}. These methods were chosen based on their 
algorithmic nature and compatibility with fine-tuning for downstream tasks. Magnitude Pruning reduces model size by removing weights 
with the smallest absolute values, while SparseGPT and Wanda leverage calibration data during the pruning process to maintain model performance.

Our methodology involved calibrating SparseGPT and Wanda with 128 random samples from the C4 dataset \cite{10.5555/3455716.3455856} to achieve 50\% sparsity. 
We measured performance metrics, including Loss and Perplexity, on 5,000 random samples from the Unnatural dataset \cite{honovich2022unnaturalinstructionstuninglanguage}. 
To ensure consistency, these samples were also used to evaluate Jensen-Shannon (JS) Divergence \cite{61115}, providing a comprehensive assessment of 
model alterations post-compression.

The Jensen-Shannon (JS) Divergence is defined as:
\begin{equation}
    \text{JS}(P \parallel Q) = \frac{1}{2} \text{KL}(P \parallel M) + \frac{1}{2} \text{KL}(Q \parallel M)
\end{equation}
where \( M = \frac{1}{2} (P + Q) \) and \(\text{KL}\) denotes the Kullback-Leibler Divergence. The KL Divergence is given by:
\begin{equation}
    \text{KL}(P \parallel Q) = \sum_{i} P(i) \log \frac{P(i)}{Q(i)}
\end{equation}

Here, \(P\) and \(Q\) are the probability distributions being compared, and \(M\) is the average of these distributions. The terms \(P(i)\) and \(Q(i)\) represent the 
probability of the \(i\)-th event in distributions \(P\) and \(Q\) respectively.

Jensen-Shannon (JS) Divergence is introduced as a crucial evaluation metric for LLM compression, offering a more nuanced understanding of how compression 
techniques impact model behavior than traditional metrics like perplexity. While perplexity focuses on next-token prediction confidence, JS Divergence 
quantifies the overall similarity between the output distributions of the original and compressed models. This makes it particularly valuable for evaluating 
methods like SparseGPT and Wanda, which aim to induce sparsity while preserving model functionality. By providing a comprehensive view of how compression 
affects the entire output distribution, JS Divergence serves as a robust measure for assessing the preservation of both general and task-specific capabilities, 
crucial aspects of successful LLM compression.

This study further examines how different types of calibration data influence compression outcomes. By comparing the efficacy of SparseGPT and Wanda using both 
general-purpose (C4 \cite{10.5555/3455716.3455856}) and task-specific (Alpaca \cite{alpaca}) calibration data, the research offers insights into how the choice of calibration data affects the performance of compressed models across various tasks.

To ensure a comprehensive and unbiased evaluation, we employ two distinct instruction-following datasets: Alpaca\footnote{The Alpaca dataset was used to fine-tune 
the original LLaMA model for instruction-following capabilities \cite{alpaca}.} for both calibration and initial 
evaluation, and Unnatural as an independent test set. This dual-dataset approach enables a robust assessment of how well the sparsified models generalize 
to novel instruction-following scenarios beyond their calibration domain. By doing so, the research not only evaluates the immediate effects of compression 
but also probes the compressed models' adaptability to different task distributions.

This methodological approach underscores the importance of carefully selecting and diversifying calibration data to enhance the performance and 
generalizability of compressed models. It also highlights the complex interplay between compression techniques, calibration data, and evaluation 
metrics in the pursuit of efficient yet capable language models.

\section{Evaluation} \label{sec:evaluation}

The LLaMA-2-7B model \cite{llama2}, is used in our experiments which was compressed with SparseGPT and Wanda algorithms calibrated with identical 128 random 
samples from the C4 dataset to achieve 50\% sparsity. Performance metrics, including Loss and Perplexity, are measured on 5,000 random samples 
from the Unnatural dataset \cite{honovich2022unnaturalinstructionstuninglanguage}, which are also used to evaluate JS Divergence, ensuring consistency in our evaluation.

\subsection{General Performance} \label{sec:gperformance}
To establish a baseline for the general performance of the base and compressed models, we evaluate them using both Cross-Entropy (Loss) 
and Perplexity, despite their close relation. Cross-Entropy is included since it was used to train and optimize the base model \cite{llama2}. 

\begin{table}[ht]
    \centering
    \caption{Perplexity and Loss of LLaMA-2-7B Base compared to 50\% compression using Magnitude, SparseGPT, and Wanda and C4 for calibration.}
    \label{tab:performance}
    \begin{tabular}{lcccccc}
        \toprule
        \multirow{2}{*}{\textbf{LLaMA-2-7B}} & \multirow{2}{*}{\textbf{Sparsity}} & \multicolumn{2}{c}{\textbf{Cross-Entropy (Loss)}} & \multicolumn{2}{c}{\textbf{Perplexity}} \\
        \cmidrule(lr){3-4} \cmidrule(lr){5-6}
        & & \textbf{Abs.} & \textbf{Rel.} & \textbf{Abs.} & \textbf{Rel.} \\
        \midrule
        Base & 0\% & 0.9814 & - & 2.6685 & - \\
        Magnitude & 50\% & 1.9021 & 93.86\% & 6.7001 & 151.12\% \\
        SparseGPT & 50\% & \textbf{0.9956} & \textbf{1.45\%} & \textbf{2.7064} & \textbf{1.42\%} \\
        Wanda & 50\% & 1.0443 & 6.41\% & 2.8416 & 6.48\% \\
        \bottomrule
    \end{tabular}
    
\end{table}

As seen in Table \ref{tab:performance}, both SparseGPT and Wanda demonstrate the ability to restore performance close to the base model when measured by Perplexity\footnote{The 
values for Perplexity differ from those reported in \cite{sun2024simpleeffectivepruningapproach} due to the choice of evaluation dataset, however, the overall trends 
are consistent.}. This would indicate that SparseGPT and Wanda are effective in compressing the LLaMA-2-7B model while maintaining its performance.

Magnitude Pruning, on the other hand, shows a significant increase in both Loss and Perplexity, suggesting a greater degradation in performance. 
This aligns with previous findings \cite{frantar2023sparsegptmassivelanguagemodels, wanda} that Magnitude Pruning, while effective in inducing sparsity, can result 
in more substantial performance drops at higher sparsity compared to methods like SparseGPT and Wanda.

\subsection{Downstream Task Performance} \label{sec:dsperformance}
We further evaluate the performance of the compressed models on their instruction-following capabilities using the Unnatural dataset, 
measuring performance with Exact Match (EM), F1 Score, and ROUGE-1 metrics. The choices of these metrics are explained in Appendix \ref{app:downstream}. 
The same 5,000 random samples from Unnatural are used to evaluate perplexity. The results are shown in Table \ref{tab:ds_performance}.

\begin{table}[ht]
    \centering
    \caption{Performance of Base compared with 50\% compression using Magnitude, SparseGPT, and Wanda on downstream tasks.}
    \label{tab:ds_performance}
    \resizebox{\textwidth}{!}{
    \begin{tabular}{lcccccccc}
        \toprule
        \multirow{2}{*}{\textbf{LLaMA-2-7B}} & \multicolumn{2}{c}{\textbf{EM}} & \multicolumn{2}{c}{\textbf{F1}} & \multicolumn{2}{c}{\textbf{ROUGE-1}} & \multicolumn{2}{c}{\textbf{Perplexity}} \\
        \cmidrule(lr){2-3} \cmidrule(lr){4-5} \cmidrule(lr){6-7} \cmidrule(lr){8-9}
        & \textbf{Abs.} & \textbf{Rel.} & \textbf{Abs.} & \textbf{Rel.} & \textbf{Abs.} & \textbf{Rel.} & \textbf{Abs.} & \textbf{Rel.} \\
        \midrule
        Base & 0.0242 & - & 0.1126 & - & 0.1190 & - & 2.6685 & - \\
        Magnitude & 0.0060 & -75.21\% & 0.0406 & -63.94\% & 0.0467 & -60.76\% & 6.7001 & 151.12\% \\
        SparseGPT & \textbf{0.0084} & \textbf{-65.29\%} & \textbf{0.0738} & \textbf{-34.46\%} & \textbf{0.0787} & \textbf{-33.87\%} & \textbf{2.7064} & \textbf{1.42\%} \\
        Wanda & 0.0038 & -84.30\% & 0.0631 & -43.96\% & 0.0684 & -42.52\% & 2.8416 & 6.48\% \\
        \bottomrule
    \end{tabular}
    }
\end{table}

SparseGPT and Wanda restore performance as measured by Perplexity (see Table \ref{tab:performance}), but exhibit degradation compared to the base model on downstream tasks. 
This indicates that Perplexity does not provide a clear picture of the true impact of compression on the model's usefulness for specific tasks \cite{jaiswal2024compressingllmstruthrarely}.

While the perplexity score is relatively low (generally a good sign), the EM, F1, and ROUGE-1 scores are all quite low. This discrepancy suggests 
that, although the model might capture the overall probability distribution of the text (as indicated by the low perplexity), it struggles with generating 
precise or highly accurate text outputs when compared to the reference texts. This highlights the limitations of using perplexity alone as a measure of model 
quality for specific downstream tasks.

\subsection{Jensen-Shannon Divergence (JS) Evaluation}\label{res:js}

\begin{figure}[ht]
    \centering
    \includegraphics[width=\linewidth]{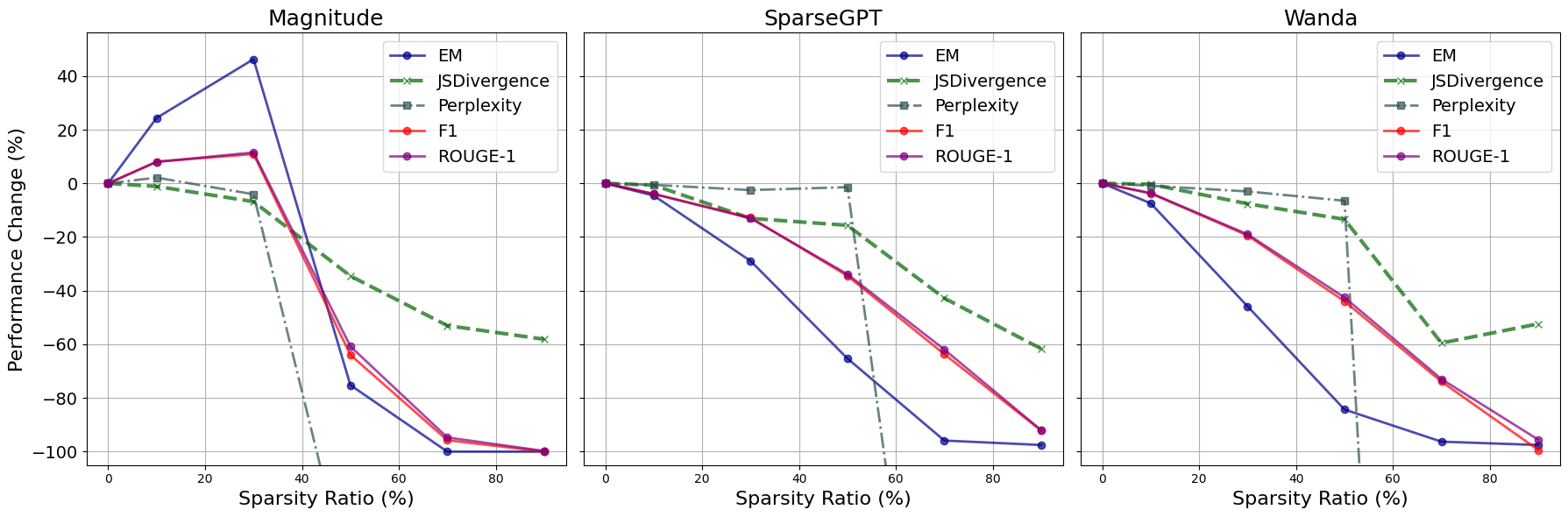}
    \caption{JS Divergence evaluated on compressed models against general and downstream task metrics.}
    \label{fig:js_divergence}
\end{figure}

We propose Jensen-Shannon (JS) Divergence \cite{61115} as a comprehensive metric to assess the impact of compression on downstream task performance.
This choice is motivated by KL Divergence's proven effectiveness in measuring model drift during inference and quantifying performance differences between student 
and teacher models \cite{xu2024surveyknowledgedistillationlarge} and the accuracy test for quantization \cite{llamacpp}. JS Divergence, being a symmetrized and bounded 
version of KL Divergence, offers additional advantages for comparing probability distributions. In our evaluation, we compare the probability distributions of outputs 
from the base model and the sparse models listed in Table \ref{tab:performance}, using the same 5,000 samples employed for Perplexity assessment. Our findings are 
shown in Figure \ref{fig:js_divergence}.

As shown in Figure 1, increasing sparsity progressively impacts JS Divergence, EM, ROUGE-1, and F-1, whereas Perplexity remains constant 
until reaching 50\% sparsity. In this context, JS Divergence more effectively captures the impact of compression and may serve as a 
superior metric for evaluating a compressed model's support for downstream tasks. Higher JS Divergence values indicate a greater 
departure from the base model's output distribution, suggesting potentially worse multitask performance. SparseGPT exhibits greater 
changes from the base model compared to Wanda up to 50\% sparsity, likely due to its error correction process during compression \cite{frantar2023sparsegptmassivelanguagemodels}. 
This process updates weights using calibration data to mitigate pruning errors, acting similarly to fine-tuning, which introduces more generalized alterations 
to the model and results in higher divergence from the base model. Beyond 50\% sparsity, results are inconclusive. It is worth noting that 
Magnitude pruning shows improved performance compared to the base model when measured by EM, ROUGE-1, and F-1 at low sparsity levels, 
likely due to the removal of redundant parameters in the model. While JS Divergence indicates changes, which may be improvements or 
degradation from the base model, these changes may indicate improvements in lower sparsity regions.

While JS Divergence captures the performance of compressed models better than Perplexity, it does not alone provide a clear picture of its superiority. 
Therefore, we further evaluate model performance using GPT-4\footnote{\texttt{gpt-4-0613} model specifically, see Appendix \ref{app:gpt4o} and 
\ref{app:gptvgpt4o} for comparison with GPT-4o.} 
as a large language model (LLM) judge. Given GPT-4's size and capabilities, it offers a human-level 
judgment and has been established to match both controlled and crowdsourced human preferences \cite{NEURIPS2023_91f18a12}. By comparing the evaluation results 
from JS Divergence to those from GPT-4, we aim to establish a more comprehensive understanding of the model's performance.

\begin{figure}[ht]
    \centering
    \includegraphics[width=\textwidth]{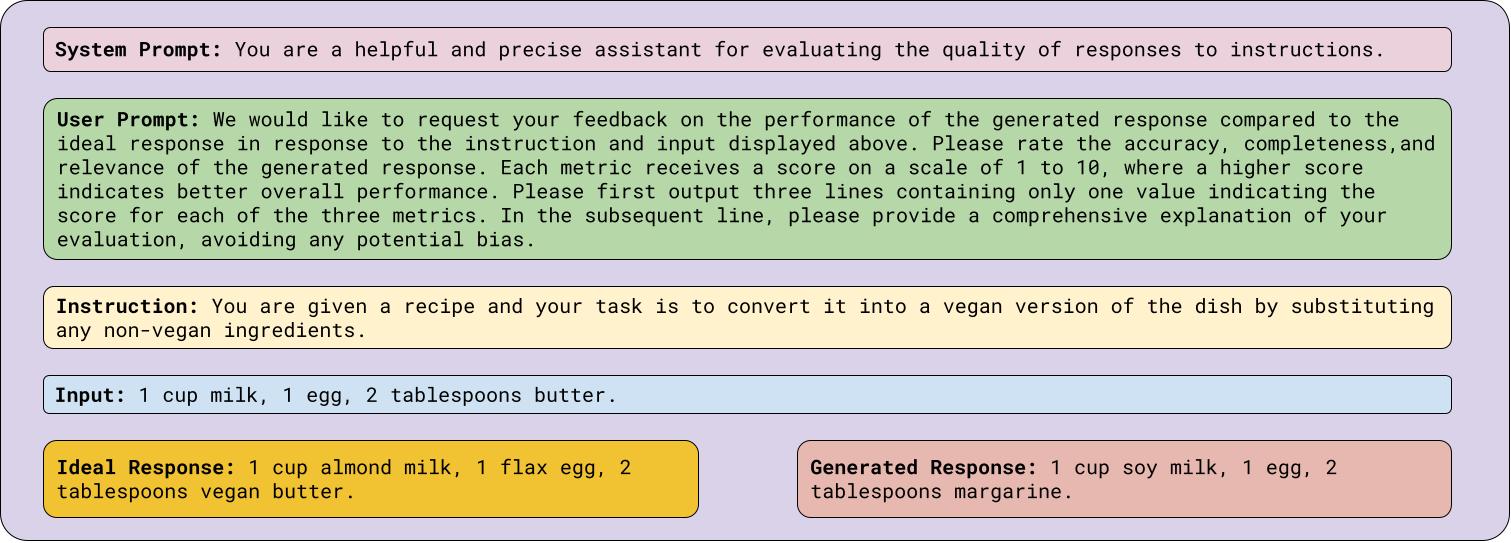}
    \caption{Template used for evaluating the quality of responses generated by the compressed models with GPT-4. 
    It includes a system prompt, user prompt, instruction-input pair, ideal response, and generated response 
    (Wanda at 30\% sparsity), providing a structured approach for assessing accuracy, completeness, and relevance.}
    \label{fig:evaluation_template}
\end{figure} 

\begin{figure}[!ht]
    \centering
    \includegraphics[width=\linewidth]{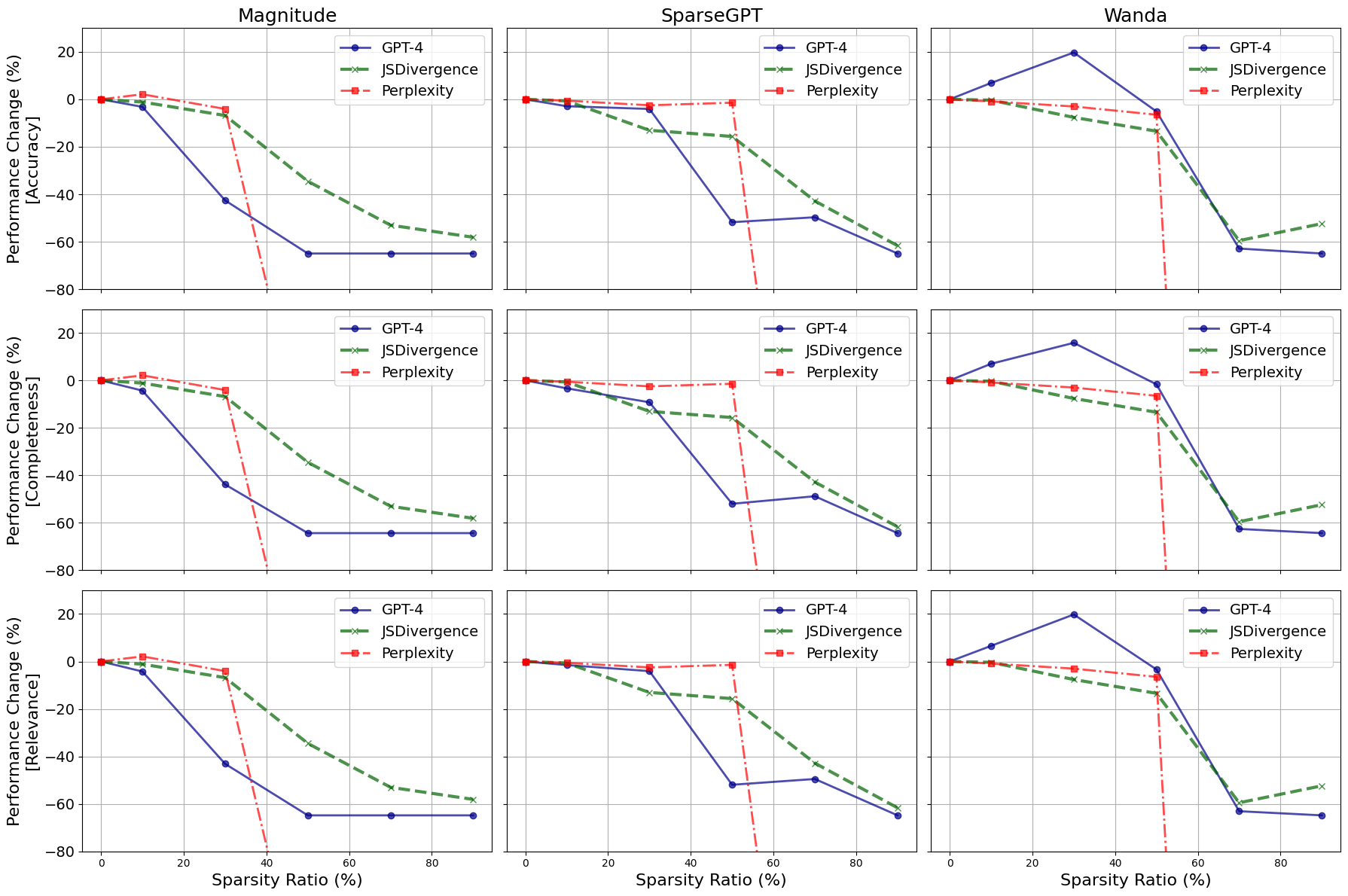}
    \caption{GPT-4 Evaluation compared with JS Divergence and Perplexity on compressed models.}
    \label{fig:gpt_4}
\end{figure}

To confirm GPT-4 as a reliable and precise evaluator, we meticulously designed prompts with clear instructions and detailed evaluation criteria, 
as shown in Figure \ref{fig:evaluation_template}. These directives set the expectation that GPT-4 would function as a meticulous and unbiased 
judge of response quality. To ensure a structured and consistent approach, we created a comprehensive template to organize the evaluation process inspired by \cite{jaiswal2024compressingllmstruthrarely}. 
The evaluator rates the generated responses on each of the three metrics and provides detailed feedback. This approach ensures that the evaluation 
is thorough, consistent, and unbiased. Figure \ref{fig:gpt_4} shows the GPT-4 evaluation on model responses compared to JS Divergence and Perplexity.  

As demonstrated in Figure \ref{fig:gpt_4}, JS Divergence captures the impact of compression on model performance more effectively compared to Perplexity. 
The GPT-4-based evaluation, which serves as a high-quality qualitative benchmark for real-world performance, shows a clear decline in model performance with 
increasing sparsity across all three compression methods (Magnitude, SparseGPT, and Wanda). JS Divergence closely follows these trends, indicating that it 
provides a reliable and comprehensive measure of the changes induced by compression.

In contrast, Perplexity remains relatively stable up to a certain sparsity level for SparseGPT and Wanda, failing to capture the full extent of performance 
degradation observed in the GPT-4 evaluations. This discrepancy highlights the limitations of using Perplexity alone as a measure of model quality, as it does 
not fully reflect the model's ability to perform specific tasks accurately.

By measuring the divergence between the probability distributions of the base model and the compressed model, JS Divergence offers a clearer and 
more objective understanding of the impact of compression. It effectively captures nuanced alterations in the model's output distribution, providing 
insights into the compressed model's robustness and its ability to maintain performance on specific tasks. Given its alignment with the GPT-4-based 
evaluations, JS Divergence emerges as a superior metric for evaluating the performance of compressed models, offering a more accurate reflection of 
real-world performance than Perplexity.

\section{Impact of Calibration Data on LLM Compression}
The impact of the number of calibration samples for model compression has been explored before \cite{jaiswal2024compressingllmstruthrarely}. 
However, we further examine the choice of calibration data on compressing LLMs by comparing the efficacy of SparseGPT and Wanda sparsification techniques using 
the Alpaca dataset for calibration, as compared to the C4 dataset used in the previous sections.

We use two distinct instruction-following datasets, Alpaca \cite{alpaca} and Unnatural \cite{honovich2022unnaturalinstructionstuninglanguage}, to evaluate the 
performance of these models. While Alpaca is used for both calibration and evaluation, providing a consistent basis for task-specific evaluation, the Unnatural \
dataset serves as an independent test set.  This dual-dataset approach allows us to assess how well the sparsified models performed on different 
instruction-following tasks, offering a robust evaluation of their capabilities within the instruction-following domain.

The results of our experiments are summarized in the following Table \ref{tab:calibration_performance}, highlighting the performance of LLaMA-2-7B and its 
sparsified variants evaluated on the Alpaca and Unnatural datasets.

\begin{table}[ht]
    \centering
    \caption{Performance of LLaMA-2-7B and its compressed variants, calibrated with C4 and Alpaca datasets, on Alpaca and Unnatural datasets.}
    \label{tab:calibration_performance}
    \resizebox{\textwidth}{!}{
    \begin{tabular}{llccccccc}
        \toprule
        \multirow{2}{*}{\textbf{LLaMA-2-7B}} & \multirow{2}{*}{\textbf{Evaluation}} & \multicolumn{2}{c}{\textbf{EM}} & \multicolumn{2}{c}{\textbf{F1}} & \multicolumn{2}{c}{\textbf{ROUGE-1}} & \multirow{2}{*}{\textbf{JS Divergence}} \\
        \cmidrule(lr){3-4} \cmidrule(lr){5-6} \cmidrule(lr){7-8}
        & & \textbf{Abs.} & \textbf{Rel.} & \textbf{Abs.} & \textbf{Rel.} & \textbf{Abs.} & \textbf{Rel.} & \\
        \midrule
        \multirow{2}{*}{Base} & Alpaca & 0.0054 & - & 0.1241 & - & 0.1376 & - & - \\
        & Unnatural & 0.0242 & - & 0.1126 & - & 0.1191 & - & - \\
        \midrule
        \multirow{2}{*}{SparseGPT-C4} & Alpaca & 0.0026 & -51.85\% & 0.1123 & -9.51\% & 0.1260 & -8.43\% & 0.2151 \\
        & Unnatural & 0.0084 & -65.29\% & 0.0738 & -34.46\% & 0.0787 & -33.92\% & 0.1431 \\
        \midrule
        \multirow{2}{*}{SparseGPT-Alpaca} & Alpaca & \textbf{0.0062} & \textbf{14.81\%} & \textbf{0.1197} & \textbf{-3.55\%} & \textbf{0.1364} & \textbf{-0.87\%} & \textbf{0.1585} \\
        & Unnatural & \textbf{0.0317} & \textbf{30.99\%} & \textbf{0.1193} & \textbf{5.95\%} & \textbf{0.1259} & \textbf{5.71\%} & \textbf{0.0794} \\
        \midrule
        \multirow{2}{*}{Wanda-C4} & Alpaca & 0.0018 & -66.67\% & 0.0904 & -27.16\% & 0.1037 & -24.64\% & 0.2206 \\
        & Unnatural & 0.0038 & -84.30\% & 0.0631 & -43.96\% & 0.0684 & -42.57\% & 0.1444 \\
        \midrule
        \multirow{2}{*}{Wanda-Alpaca} & Alpaca & \textbf{0.0042} & \textbf{-22.22\%} & \textbf{0.1008} & \textbf{-18.78\%} & \textbf{0.1128} & \textbf{-18.02\%} & \textbf{0.1912} \\
        & Unnatural & \textbf{0.0090} & \textbf{-62.81\%} & \textbf{0.0736} & \textbf{-34.64\%} & \textbf{0.0784} & \textbf{-34.17\%} & \textbf{0.1357} \\
        \bottomrule
    \end{tabular}
    }
\end{table}

Our experiments reveal that calibration data significantly influences the effectiveness of model compression, with models calibrated using the Alpaca 
dataset generally outperforming those calibrated with C4 across various metrics. SparseGPT shows higher sensitivity to the choice of calibration data 
than Wanda, as SparseGPT-Alpaca models retain or slightly improve performance in some metrics, while C4-calibrated models exhibit significant performance 
declines. Wanda models also degrade with C4 calibration, but less severely.

SparseGPT-Alpaca achieves the most impressive results, particularly in the Unnatural dataset evaluation, indicating that task-specific calibration not 
only preserves in-domain capabilities but also enhances the model's ability to generalize to novel instruction-following scenarios. The consistently 
strong performance of Alpaca-calibrated models across both evaluation datasets underscores the value of using task-specific datasets for calibration.

\section{Discussion} \label{sec:discussion}

This study evaluates the impact of three popular compression techniques — Magnitude Pruning, SparseGPT, and Wanda—on the LLaMA-2-7B model. 
The key findings reveal several critical insights into the trade-offs and effectiveness of these methods in preserving model performance while reducing complexity.

The results indicate that while SparseGPT and Wanda maintain perplexity levels close to the base model, they exhibit significant degradation 
in downstream task performance. This disparity underscores the inadequacy of perplexity as a sole evaluation metric for assessing the efficacy 
of compression techniques. Perplexity measures how confidently a model predicts the next token, but it does not fully capture the nuanced impacts 
of compression on task-specific outputs. Magnitude Pruning, although effective in inducing sparsity, showed a notable increase in both Loss and Perplexity, 
suggesting a greater degradation in overall performance. However, it was observed that at low sparsity levels, Magnitude Pruning could improve 
downstream task performance. This phenomenon is likely due to the removal of redundant parameters, which may improve the model for specific tasks.

Although previous studies \cite{jaiswal2024compressingllmstruthrarely,Muhlgay2023GeneratingBF,hu2024can,sourabh2024ppl} have highlighted the problems of 
using perplexity as the sole evaluation metric for LLM compression, they fail to suggest an alternative. This study addresses this gap by proposing 
Jensen-Shannon (JS) Divergence as a more comprehensive metric, offering deeper insights into model changes post-compression. Our results reveal that, 
unlike perplexity, which remains constant up to 50\% sparsity, JS Divergence effectively captures the impact of compression on downstream task performance, 
indicating greater changes in the model's output distribution. SparseGPT exhibited higher JS Divergence compared to Wanda, indicating more generalized 
changes due to its error correction process. This process, which updates weights using calibration data, acts similarly to fine-tuning and introduces more 
extensive alterations to the model. In contrast, Wanda's lower JS Divergence suggests closer adherence to the base model but correlates with poorer downstream 
task performance compared to SparseGPT.

The integration of GPT-4 as an evaluator in this study serves as an effective proxy for real-world task performance. GPT-4's assessments closely mirror 
human judgment \cite{NEURIPS2023_91f18a12}, offering valuable insights into the practical implications of compressing language models. The strong alignment between GPT-4 evaluations 
and JS Divergence further validates JS Divergence as a reliable metric for assessing compressed model performance. Although evaluation through GPT-4 reflects 
real-world performance, it is time consuming and expensive for large-scale experiments. Hence, the alignment with JS Divergence suggests that JS can be used 
as a more practical and cost-effective alternative for evaluating compressed models.

The choice of calibration data significantly influences the effectiveness of model compression. Task-specific calibration data, such as from the 
Alpaca dataset, significantly enhances the performance of compressed models on downstream tasks compared to general calibration data like C4. 
SparseGPT, in particular, demonstrates higher sensitivity to the choice of calibration data, retaining or even improving performance with 
task-specific calibration, while showing significant drops with general calibration data. SparseGPT and Wanda show varying sensitivity to 
calibration data. SparseGPT calibrated with Alpaca generally retains or improves performance on some metrics, whereas SparseGPT calibrated 
with C4 shows significant performance drops. Wanda, while effective, shows more sensitivity to calibration data but retains relatively better 
results when calibrated with task-specific data.

\section{Conclusion} \label{sec:conclusion}

Our findings highlight the limitations of using perplexity as the sole evaluation metric for LLM compression. While compression methods may preserve the perplexity 
of the original model, they often result in significant performance declines on specific downstream tasks. This underscores the need for a more comprehensive evaluation approach that 
captures the nuanced effects of compression. Jensen-Shannon (JS) Divergence emerges as a valuable tool in this context, providing deeper insights into the trade-offs 
between model size and task-specific capabilities. The strong alignment between GPT-4 evaluations and JS Divergence further validates JS Divergence as a comprehensive 
metric for assessing the real-world impact of compression. Additionally, our experiments emphasize the critical role of calibration data in successful LLM sparsification, 
showing that task-specific calibration data significantly enhances the performance of compressed models on downstream tasks.

A key area for future research is the integration of fine-tuning with compression to optimize performance and complexity for downstream tasks. 
Our future study aims to address this gap by investigating how fine-tuning can be effectively combined with compression methods to enhance 
task-specific performance while maintaining model efficiency. Since the most effective compression techniques leverage calibration data, integrating 
fine-tuning with compression holds significant potential. Fine-tuning allows models to adapt to specific tasks, and when combined with compression 
techniques that utilize task-specific calibration data, it can enhance the overall efficacy and robustness of the model. Future research should 
explore the synergistic effects of fine-tuning and compression, particularly how calibration data can be leveraged during the fine-tuning process 
to optimize both performance and complexity.

In conclusion, while SparseGPT and Wanda show promise for compressing LLMs, addressing performance gaps on downstream tasks remains a challenge. Our study 
advocates for using metrics like JS Divergence alongside perplexity to better evaluate compression techniques. This approach can help develop more effective 
compression methods, enabling the use of powerful LLMs in resource-constrained environments without losing their practical utility. By adopting comprehensive 
evaluation metrics, researchers can better understand how compression affects the practical use of large language models. This will aid in the adoption of LLMs 
for specific tasks, highlighting the practical value of proper compression methods for efficient and effective use in various domains.

\section*{Acknowledgements}
The authors would like to express their appreciation to the entire team at OptiML.org, whose support and constructive feedback contributed 
significantly to the refinement of this research. We also extend our gratitude to our colleagues for their thorough review and insightful 
comments on the draft.

\newpage
\nocite{*}
\printbibliography

% \appendix
% \numberwithin{equation}{section}
\newpage
\appendix

\section{Downstream Task Metrics}\label{app:downstream}
When evaluating instruction-following tasks for language models, it is crucial to select appropriate metrics that accurately reflect 
the model's performance. The chosen metrics—Exact Match (EM), F1 Score, and ROUGE-1 F1 Score—each offer unique insights into different 
aspects of the model's output quality.

It's worth noting that ROUGE-1 refers to ROUGE-1 F1, which is the harmonic mean of precision and recall, measuring "intersecting" tokens. 
It is typically denoted as ROUGE-1. The F1 Score, in the context of instruction-following tasks for language models, is usually based 
on exact matches between the generated text and the reference.

The key distinction between these metrics lies in their strictness and how they handle word matching:

\begin{itemize}
    \item Exact Match (EM): This is the strictest metric, requiring the generated text to match the reference exactly, word for word. 
    It is particularly useful for tasks where precision is paramount, and any deviation from the reference text is considered an error. 
    
    \item F1 Score: This metric is typically calculated based on the overlap of exact words or subwords between the generated text 
    and the reference. It's less strict than EM as it allows for partial matches, but it still requires words to match exactly. 
    This metric balances precision and recall, making it suitable for tasks where both exact matches and partial matches are important. 
    The F1 score is particularly relevant for instruction-following tasks as it allows for partial matches, rewarding the model for 
    correctly predicting some of the words even if the entire sequence is not perfectly matched.
    \item ROUGE-1 F1 Score: This is the most flexible of the three metrics. It's order-independent, focusing on the presence of the same 
    words regardless of their sequence. The model output receives credit for including the same words as the reference, even if they 
    appear in a different order. This metric is useful for tasks where the overall content and meaning are more important than the exact 
    word order. The ROUGE-1 score is calculated by evaluating the overlap of unigrams between the prediction and reference.
\end{itemize}

These metrics are chosen for instruction-following benchmarks because they collectively provide a comprehensive evaluation of the 
model's performance. The Exact Match metric ensures that the model can produce precise outputs when necessary. The F1 Score offers 
a balanced view of the model's ability to generate both exact and partial matches. Finally, the ROUGE-1 F1 Score allows for flexibility 
in word order, which is often important in natural language generation tasks.

\section{GPT-4o Evaluation}\label{app:gpt4o}

\begin{figure}[!ht]
    \centering
    \includegraphics[width=\linewidth]{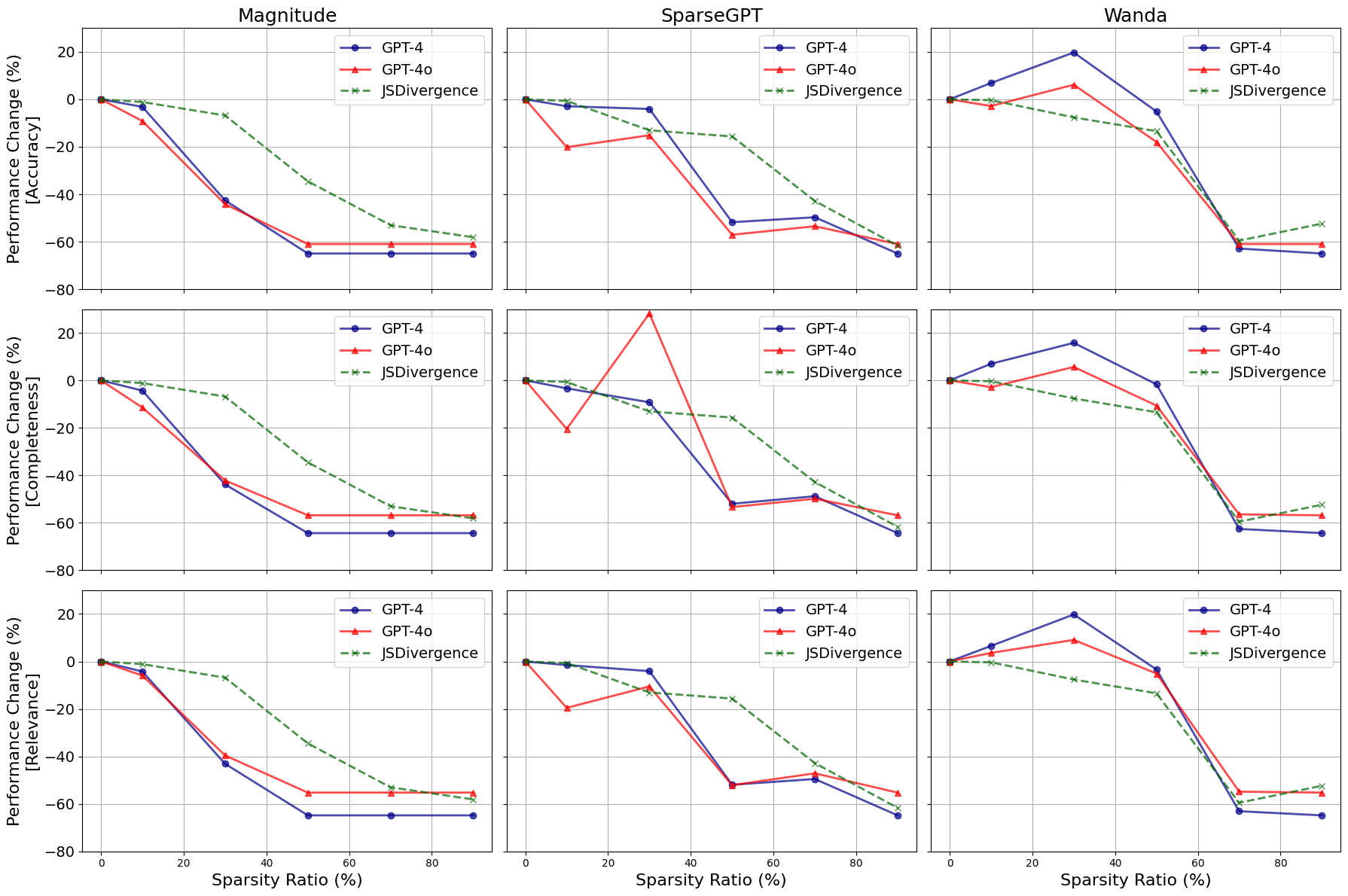}
    \caption{GPT-4 vs GPT-4o evaluation compared with JS Divergence on compressed models.}
    \label{fig:gpt_4o}
\end{figure}

As the latest iteration in the GPT series, GPT-4 omni (GPT-4o) has significantly improved its capabilities, enhancing both understanding 
and generation. Hence, we also evaluate the compression model performance with GPT-4o similarly to GPT-4. Figure \ref{fig:gpt_4o} shows the 
comparison of evaluation done with JS Divergence against two different models of GPT-4. 

From the plots, we observe that both GPT-4 and GPT-4o exhibit similar trends in performance degradation as the sparsity ratio increases. 
JS Divergence increases with the sparsity ratio for all compression methods, indicating a higher divergence between the compressed and 
original models. This trend is consistent across both GPT-4 and GPT-4o evaluations. The consistency in the evaluations of these three metrics 
suggests that either model can be effectively used for evaluating the performance of LLM compression.

\section{GPT-4 vs GPT-4o}\label{app:gptvgpt4o}

While evaluating the performance of response generation for different sparsity ratios in Magnitude Pruning, SparseGPT, and Wanda, we also 
compared GPT-4 and GPT-4o. Given the same instruction-input pairs, both models generated responses, which were then evaluated by both GPT-4 
and GPT-4o as explained in Section \ref{res:js}. Table \ref{tab:gpt4_vs_gpt4o} shows the performance of these responses when evaluated by 
each model.

\begin{table}[!ht]
    \centering
    \resizebox{\textwidth}{!}{
    \begin{tabular}{lcccccc}
        \toprule
        \multirow{2}{*}{\textbf{LLM}} & \multicolumn{3}{c}{\textbf{GPT-4 Evaluation (1 - 10)}} & \multicolumn{3}{c}{\textbf{GPT-4o Evaluation (1 - 10)}} \\
        \cmidrule(lr){2-4} \cmidrule(lr){5-7}
         & \textbf{Accuracy} & \textbf{Completeness} & \textbf{Relevance}  & \textbf{Accuracy} & \textbf{Completeness} & \textbf{Relevance} \\
        \midrule
        \textbf{GPT-4} & \textbf{8.3137} & \textbf{8.7941} & \textbf{8.8431}  & 7.2857 & 7.3452 & 7.6071  \\
        \textbf{GPT-4o} & 8.2705 & 8.5176 & 8.6823 & \textbf{8.2967} & \textbf{8.5666} & \textbf{8.7221}  \\
        \bottomrule
    \end{tabular}}
    \caption{GPT-4 vs GPT-4o evaluation of responses generated by each model.}
    \label{tab:gpt4_vs_gpt4o}
\end{table}

The results reveal a bias in the evaluations: each model rates its responses higher than those of its counterpart.
\end{document}